\documentclass[runningheads]{llncs}
\usepackage{cite}
\usepackage{times}
\usepackage{latexsym}
\usepackage{microtype}
\usepackage{graphicx}
\usepackage{hyperref}
\usepackage{multirow}
\usepackage{booktabs}
\usepackage{nicefrac}
\usepackage{subfigure}
\usepackage{color}
\usepackage{amsmath,amsfonts,amssymb}
\graphicspath{{./imgs/}}

\begin{document}
\title{Label-Wise Document Pre-Training for Multi-Label Text Classification}
%
%

\author{Han Liu \and
Caixia Yuan \and
Xiaojie Wang}

\authorrunning{H. Liu et al.}

\institute{Center of Intelligence Science and Technology, \\
Beijing University of Posts and Telecommunications, China
\email{\{liuhan,yuancx,xjwang\}@bupt.edu.cn}}

\maketitle              
\begin{abstract}
A major challenge of multi-label text classification (MLTC) is to stimulatingly exploit possible label differences and label correlations. In this paper, we tackle this challenge by developing \textbf{L}abel-\textbf{W}ise \textbf{P}re-\textbf{T}raining (\textbf{LW-PT}) method to get a document representation with label-aware information. The basic idea is that, a multi-label document can be represented as a combination of multiple label-wise representations, and that, correlated labels always cooccur in the same or similar documents. LW-PT implements this idea by constructing label-wise document classification tasks and trains label-wise document encoders. Finally, the pre-trained label-wise encoder is fine-tuned with the downstream MLTC task. Extensive experimental results validate that the proposed method has significant advantages over the previous state-of-the-art models and is able to discover reasonable label relationship. The code is released to facilitate other researchers.
\footnote{\url{https://github.com/laddie132/LW-PT}}

\end{abstract}

\section{Introduction}
Multi-label text classification (MLTC) is a task of assigning a document into one or more class labels. In recent years, MLTC have been widely used in many scenarios, such as tag recommendation\cite{furnkranz2008multilabel}, information retrieval\cite{gopal2010multilabel}, sentiment analysis\cite{cambria2014senticnet}, and so on. Different from traditional single label classification, where one instance is associated with one target label, in MLTC, a text is naturally associated with multiple labels. This makes it both an essential and challenging task in Natural Language Processing (NLP).

A straightforward solution to MLTC is to decompose the problem into a series of binary text classification problems \cite{boutell2004learning}, each for one label. Such a solution, however, neglects the fact that information of one label may be helpful for learning another related label, whereas in real-world multi-label text classification applications, one label might be associated with other labels. For example, in music tagging task, music style ``metal'' is naturally associated with ``rock''. It is well-accepted that, in order to achieve a good performance, we should not only consider the difference between labels, but also correlation. Especially when some labels have insufficient training examples, the label correlations may provide helpful extra information.

In recent years, deep learning models such as CNN \cite{kim2014convolutional} and LSTM \cite{hochreiter1997long} have been firmly established as state-of-the-art approaches in text representation. However, these methods cannot learn a label difference since they simply use logistic regression for each label independently to achieve multi-label classification. There are also sequence learning models that attempt for MLTC, such as CNN-RNN \cite{chen2017ensemble}, and Sequence Generation Model (SGM)\cite{yang2018sgm}, which generate a sequence of possible labels using a RNN decoder. Although the label difference can be captured by attention on the state of decoder, correlation of any two labels cannot be dynamically modeled through the fixed recurrent decoder. 

To alleviate the above existing problems, we proposed a novel pre-training task and model \textit{Label-Wise Pre-Training (LW-PT)} to get label-wise document representation. Given a target document and one label it possesses, we sample a document as a positive example which also has this label, and several documents as negative ones that don't take this label. The pre-training goal is to learn a classifier distinguishing between positive and negative examples. In this way, we can train a document encoder that captures label-sensitive information. Finally, each document is represented as a concatenation of all label-wise representations.

Specificly, to explicitly model the label difference, we propose two label-wise encoders by self-attention mechanism into the pre-training task, including \textit{Label-Wise LSTM (LW-LSTM)} encoder for short documents and \textit{Hierarchical Label-Wise LSTM (HLW-LSTM)} for long documents. For document representation on each label, they share the same LSTM layer and use different attention weights for different labels.

Obviously, labels appeared together in a document may have similar concepts. For example, music style label ``metal'' and ``rock'' tend to appear together, thus pre-training tasks for these two labels may share the same or similar training samples. Therefore, such label-wise document representation can also capture the label correlations.

The label-wise document representation is fine-tuned with a MLP layer for multi-label classification. Experiments demonstrate that our method achieves a state-of-art performance and has a substantial improvement compared with several strong baselinses.

Our contributions are as follows:
\begin{enumerate}
    \item We propose a novel label-wise pre-training task and model LW-PT for MLTC task, which encodes document with a combination of label-wise representations by effectively exploiting both label difference and correlation.
    \item Two label-wise encoder LW-LSTM and HLW-LSTM are proposed for pre-training and fine-tuning, capturing label-aware information for both short and long document.   
    \item To the best of our knowledge, this is the first pre-training task to obtain a label-wise document representation. Experiments show that the proposed method achieves outstanding performance than previous approaches on several datasets.
\end{enumerate}

\section{Method}
\begin{figure}[htb]
    \centering
    \subfigure[Pre-Training]{
        \includegraphics[width=0.4\textwidth]{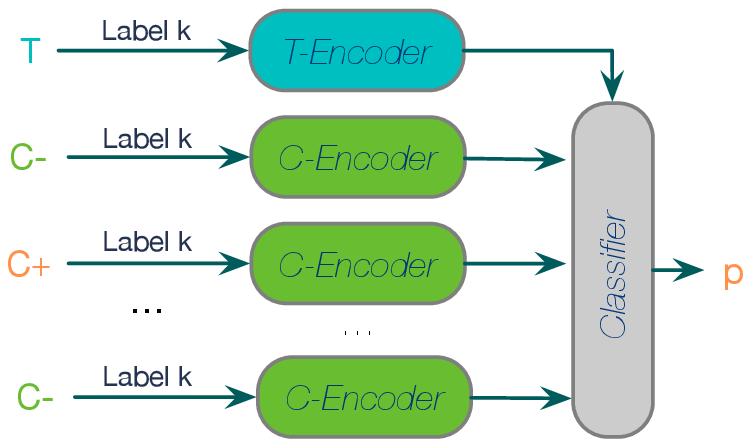}
    }
    \subfigure[Task-MLTC]{
        \includegraphics[width=0.4\textwidth]{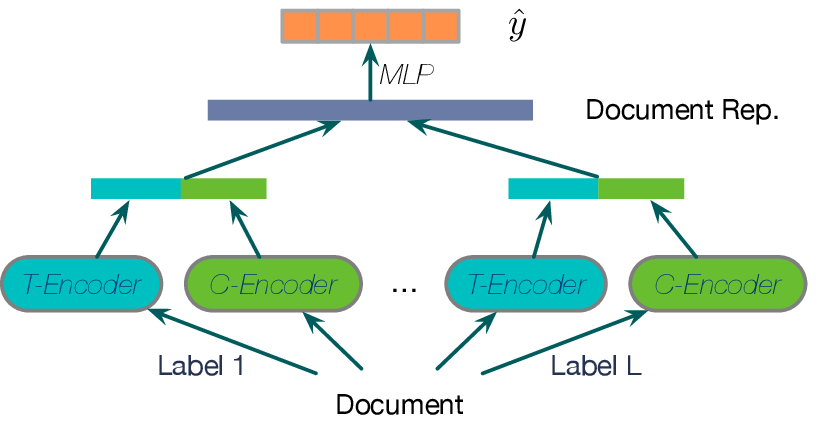}
    }
    \caption{An overview of the proposed approach. (a) is the pre-training task and model LW-PT. (b) is the downstream MLTC model.}
    \label{model:overview}
\end{figure}

This section introduces the proposed method, including pre-trained task and multi-label classification model, as shown in the Figure \ref{model:overview}. The former constructs a label-wise representation for document, and the latter fuses the multi-label representation for MLTC.

\subsection{Task \& Model}
We design a multi-label pre-training task and model \textit{Label-Wise Pre-Training (LW-PT)}. When given a target document $T$ and it's label set $\{1,...,k,...,l\}$, we sample a positive example $C^{+}$ which has the same label $k$ as target and several negative examples $C^{-}$ which doesn't have the label $k$. This two parts are combined as candidates $\{C_i\}_{i=1}^n$ with size $n$. The training goal is to discriminate the positive example from all $\{C_i\}_{i=1}^n$. We do this for each label $k$ in the label set of targer document to make training samples.

Figure \ref{model:overview}-a shows the architecture of pre-training task and model LW-PT. We denote the target encoder as \texttt{T-Encoder} and candidates encoder as \texttt{C-Encoder}. They have different parameters. 

 Therefore, as for the label $k$, we calculate the representation for the target document $T$ and candidates $\{C_i\}_{i=1}^n$, and obtain $Q^{t}_k \in \mathbb{R}^{2d}$ and $Q^{c_i}_k \in \mathbb{R}^{2d}$.
\begin{align}
    Q^{t}_k &= \operatorname{T-Encoder}(T, k) \\
    Q^{c_i}_k &= \operatorname{C-Encoder}(C_i, k)
\end{align}

Further, the similarity of any two documents is directly calculated by dot product and trained using the negative log-likelihood loss function.
\begin{equation}
    \mathcal{L}_{pt} = \mathop{\mathbb{E}}\limits_{t,k,c^+,c^-}\left[-\log \frac{\exp{((Q^{t}_k)^T Q^{c^+}_k})}{\exp{((Q^{t}_k)^T Q^{c^+}_k)} + \sum_{c^-}\exp{((Q^{t}_k)^T Q^{c^-}_k})}\right]
\end{equation}

Obviously, we can choose different encoder for it. In MLTC task, different labels may attend on differnt part of the document. Therefore, we design a label-wise encoder. Normally, \textit{Label-Wise LSTM (LW-LSTM)} can meet the requirements. However, LSTM is not effective for long documents. Inspired by Hierarchical Attention Network (HAN) \cite{yang2016hierarchical}, we also introduce \textit{Hierarchical Label-Wise LSTM(HLW-LSTM)} to model long documents with several sentences. The details are described in the following.

\subsubsection{LW-LSTM}
We use BiLSTM and self-attention mechanism to get document representation $Q_k \in \mathbb{R}^{2d}$ on label $k$. Denote the document word-level embedding as $D\in \mathbb{R}^{t \times d}$, where $t$ is the maximum number of words in a document, $d$ is the size of embedding dimision.
\begin{align}
    H &= \operatorname{BiLSTM}(D) \\
    \alpha_k & = \operatorname{softmax}(H U_k) \\
    Q_k &= (\alpha_k)^T H
\end{align}
where $H \in \mathbb{R}^{t \times 2d}$, $\alpha_k \in \mathbb{R}^{t \times 1}$, and $U_k \in \mathbb{R}^{2d \times 1}$ is a trainable parameter for label $k$.

\subsubsection{HLW-LSTM}
Similar as Hierarchical Attention Networks (HAN) \cite{yang2016hierarchical}, we use BiLSTM and hierarchical attention mechanism. As for the label $k$ and document word-level embedding $D \in \mathbb{R}^{m \times t \times d}$, where $m$ is the maximum number of sentences in a document, $t$ is the maximum number of words in a sentence, $d$ is the size of embedding dimision. Thus, $D_{j} \in \mathbb{R}^{t \times d}$ represents the $j$-th sentence for a given document $D$.
\begin{equation}
    H^w_j = \operatorname{BiLSTM}(D_j)
\end{equation}
where $H^w_j\in \mathbb{R}^{t \times 2d}$.

After that, we obtain the $j$-th sentence vector $[S_{k}]_j \in \mathbb{R}^{2d}$ on a specific label $k$.
\begin{align}
    [\alpha^w_{k}]_j & = \operatorname{softmax}(H^w_{j} U^w_k) \\
    [S_{k}]_j &= ([\alpha^w_{k}]_j)^T H^w_{j}
\end{align}
where $[\alpha^w_{k}]_j \in \mathbb{R}^{t \times 1}$. And $U^w_k \in \mathbb{R}^{2d \times 1}$ is the word-level context vector on label $k$, which is used to measure the importance of each word on current label. This is a trainable parameter and randomly initialized.

Thus, each sentence vector $S_k \in \mathbb{R}^{m \times 2d}$ for document $D$ is obtained. Same as above, we use BiLSTM and self-attention to get document representation $Q_k \in \mathbb{R}^{2d}$.
\begin{align}
    H^s_k &= \operatorname{BiLSTM}(S_k) \\
    \alpha^s_k & = \operatorname{softmax}(H^s_k U^s_k) \\
    Q_k &= (\alpha^s_k)^T H^s_k
\end{align}
where $H^s_k \in \mathbb{R}^{m \times 2d}$, $\alpha^s_k \in \mathbb{R}^{m \times 1}$. And $U^s_k \in \mathbb{R}^{2d \times 1}$ is the sentence-level context vector on label $k$, which is used to measure the importance of each sentence on current label. This is a trainable parameter and randomly initialized.

\subsection{Multi-Label Classification}
After pre-training, we use \texttt{T-Encoder} and \texttt{C-Encoder} in the downstream model for MLTC task. As illustrated in Figure \ref{model:overview}-b, we concatenate the output of \texttt{T-Encoder} and \texttt{C-Encoder} to get document representation for each label $k$. Then, each label-wise representation is also concatenated. And we obtain $M \in \mathbb{R}^{l \times 4d}$ as the final representation for a given document $D$.
\begin{align}
    Q^{t}_{k} &= \operatorname{T-Encoder}(D, k) \\
    Q^{c}_{k} &= \operatorname{C-Encoder}(D, k) \\
    M_{k} &= [Q^{t}_{k};Q^{c}_{k}]
\end{align}

After that, we simplely use a MLP layer to predict the probability of each label $\hat{y}$.
\begin{equation}
    \hat{y} = \operatorname{sigmoid}(W^T M)
\end{equation}
where $W \in \mathbb{R}^{(l \times 4d) \times l}$ is a trainable parameter.

We use the cross-entropy loss function for MLTC task.
\begin{equation}
    \mathcal{L}_{cls} = \mathbb{E}_D\left[-\sum_{k=1}^{l}(y_{k}\log \hat{y}_{k} + (1 - y_{k}) \log (1-\hat{y}_{k}))\right]
\end{equation}
where $\hat{y}_{k}$ is the predict probability of label $k$ for current document. And $y_{k}$ is the ground truth of whether label $k$ appeared in current document.

It should be noted that \texttt{T-Encoder} and \texttt{C-Encoder} are fine-tuned in the MLTC task at the same time.

\section{Experiments}
In this section, we evaluate the proposed method on two datasets with short or long documents, and also compare with the previous state-of-art models on several widely used metrics.

\subsection{Dataset}
We use two MLTC datasets RMSC\cite{zhao2018review} and AAPD\cite{yang2018sgm} with different length and language of documents. The former has longer documents with several sentences in Chinese. And the latter is shorter in English. Table \ref{tab:dataset} shows some statistics.
\begin{itemize}
    \item \textbf{RMSC-V2}\cite{zhao2018review}: The dataset is collected from a popular Chinese music review website\footnote{\url{https://music.douban.com}}. For each music, it includes a set of human annotated styles, and associated reviews. 22 styles are defined in it. The dataset contains over 7.1k samples, 288K reviews, and 3.6M words. However, the initial version didn't contains a certain split. Thus, we split the dataset into trian/valid/test by the same ratio as the original paper.
    \item \textbf{AAPD}\cite{yang2018sgm}: The dataset is collected from a English academic website\footnote{\url{https://arxiv.org}} in the computer science field. Each sample contains an abstract and the corresponding subjects. 54 labels are defined and 55,840 samples are included.
\end{itemize}
\begin{table}[htb]
    \centering
    \setlength{\abovecaptionskip}{0.3cm}
    \begin{tabular}{c|cccccc}
        \hline
        Datasets & $N$ & $V$ & $M$ & $L$ & $\overline{L}$ & $\overline{W}$ \\ \hline
        RMSC-V2 & 5020 & 646 & 1506 & 22 & 2.22 & 497.09 \\
        AAPD & 53,840 & 1,000 & 1,000 & 54 & 2.41 & 167.27 \\
        \hline
    \end{tabular}
    \caption{Statistics of datasets. N, V, M are the number of training, validing or testing samples. L is the size of label set. $\overline{L}$ is the average number of labels per sample. $\overline{W}$ is the average number of words per document.}
    \label{tab:dataset}
\end{table}

\subsection{Evalution Metrics}
We use four widely used evaluation metrics, the same as \cite{chen2017ensemble,yang2018sgm,zhao2018review}.
\begin{itemize}
    \item \textbf{One-Error}: One-error calculates the fraction of samples whose top-1 predicted label is not in the ground truth.
    \item \textbf{Hamming Loss}: Hamming loss calculates the fraction of the wrong labels to the total number of ground truth labels.
    \item \textbf{Macro-F1}: Macro F1 takes the average of each label's F1 score.
    \item \textbf{Micro-F1}: Micro F1 calculates the F1 score over all sample-label pairs.
\end{itemize}

\subsection{Details}
For RMSC-V2 dataset, we use jieba \footnote{\url{https://github.com/fxsjy/jieba}} toolkit tokenizer and train word embedding by Skip-gram model\cite{mikolov2013distributed}. The embedding dimision and hidden size is 100. For AAPD dataset, we also train word embedding by Skip-gram model\cite{mikolov2013distributed}. The embedding dimision and hidden size is 256. 

We use 2 layer LSTM for LW-LSTM encoder. And a hierarchical LSTM layer for HLW-LSTM encoder. In addition, Adam\cite{kingma2014adam} optimizer with learning rate 0.001 is used. We add layer normalization\cite{ba2016layer} and dropout with probability 0.2. In pre-training procedure, 3k iterations are runned and the number of candidates documents is 3. Batch size is 128. In fine-tuning, 20 epochs are runned.

It should be noted that the pre-training procedure is on dataset of train. 

\subsection{Baselines}
We use several MLTC baselines to compare with our method.
\begin{itemize}
    \item \textbf{Binary Relevance (BR)}\cite{boutell2004learning} transforms the multi-label problem into several single-label classification problems and ignore the correlation between labels.
    \item \textbf{Classifier Chains (CC)}\cite{read2011classifier} transforms the multi-label problem into a chain of several single-label classification problems.
    \item \textbf{Label Powerset (LP)}\cite{tsoumakas2007multi} transforms the multi-label problem to a multi-class problem with one multi-class classifier trained on all unique label.
    \item \textbf{CNN}\cite{kim2014convolutional} attempts to use a convolutional neural network with several different size of kernels, and predicts each label with logistic regression.
    \item \textbf{LSTM}\cite{hochreiter1997long} uses a 2-layer LSTM neural network with self-attention on the last layer to get document representation, and predicts each label with logistic regression.
    \item \textbf{Hierarchical Attention Network (HAN)}\cite{yang2016hierarchical} uses a hierarchical attention network on word and sentence level to get document representation, and predicts each label with logistic regression.
    \item \textbf{HAN+LG}\cite{zhao2018review} introduces a label graph matrix into HAN.
    \item \textbf{CNN-RNN}\cite{chen2017ensemble} utilizes CNN and RNN on Seq2Seq framework to get labels one by one.
    \item \textbf{SGM+GE}\cite{yang2018sgm} computes a weighted global embedding based on all labels as opposed to just the top one at each timestep.
    \item \textbf{set-RNN}\cite{qin2019adapting} presents an adaptation of RNN sequence models to the problem, where the target is a set of labels, not a sequence.
    \item \textbf{reg-LSTM}\cite{adhikari2019rethinking} proposes a simple BiLSTM architecture with regularization.
    \item \textbf{MAGNET}\cite{pal2020multi} uses a graph attention network to capture the attentive dependency structure among the labels.
\end{itemize}

\subsection{Results}
The complete results on RMSC-V2 and AAPD datasets are shown in Table \ref{tab:results}. We calculate the One-Error, Hamming-Loss, Macro F1 score and Micro F1 score for several models. Compared to the previous approaches, our LW-LSTM encoder or HLW-LSTM encoder with pre-training mechanism presents a outstanding performance. Adding the fine-tuning procedure further improve the performance on a substantial margin.

It should be noted that LW-LSTM model denotes directly use the LW-LSTM encoder in the downstream task model without pre-training mechanism. And the same as HLW-LSTM model. We can see that pre-training and fine-tuning have a significant contribution to the performance. But for which label-wise encoder to use, it's determined by the dataset. For long documents with several sentences (e.g. RMSC), we suppose to choose HLW-LSTM encoder with hierarchical attention. And for short documents with less sentences (e.g. AAPD), we suppose to use LW-LSTM encoder. 

We can see that the Macro F1 score has an outstanding improvement when pre-training mechanism is used. This is because the pre-training approach improves the learning of labels with less frequency by capturing the correlation between labels. And Macro F1 score is the average of each label's F1 score. These experimental results is consistent with our ideas.
\begin{table}[htb]
    \centering
    \setlength{\abovecaptionskip}{0.3cm}
    \begin{tabular}{c|cccc|cccc}
        \hline
        \multirow{2}*{\textbf{Models}} & \multicolumn{4}{c|}{RMSC-V2} & \multicolumn{4}{c}{AAPD} \\ \cline{2-9}
        & \textbf{OE(-)} & \textbf{HL(-)} & \textbf{MacroF1(+)} & \textbf{MicroF1(+)} & \textbf{OE(-)} & \textbf{HL(-)} & \textbf{MacroF1(+)} & \textbf{MicroF1(+)} \\ \hline
        BR & 74.4 & 0.083 & 24.7 & 41.8 & - & 0.0316 & - & 64.6 \\
        CC & 67.5 & 0.107 & 29.9 & 44.3 & - & 0.0306 & - & 65.4 \\
        LP & 56.2 & 0.096 & 37.7 & 50.3 & - & 0.0312 & - & 63.4 \\
        CNN & 23.84 & 0.0702 & 41.11 & 59.10 & 19.90 & 0.0264 & 40.37 & 65.00 \\
        LSTM & 24.24 & 0.0688 & 36.16 & 58.81 & 18.50 & 0.0253 & 39.16 & 67.08 \\ 
        HAN & 18.26 & 0.0590 & 53.18 & 66.75 & 15.50 & 0.0236 & 51.80 & 70.81 \\
        HAN+LG & 17.60 & 0.0580 & 55.20 & 68.21 & 14.50 & 0.0235 & 52.97 & 71.19 \\ 
        CNN-RNN & - & - & - & - & - & 0.0278 & - & 66.4 \\
        SGM+GE & - & - & - & - & - & 0.0245 & - & 71.0 \\ 
        set-RNN & - & - & - & - & - & 0.0241 & 54.8 & 72.0 \\ 
        reg-LSTM & - & - & - & - & - & - & - & 70.5 \\
        MAGNET & - & - & - & - & - & 0.0252 & - & 69.6 \\ \hline
        \textbf{LW-LSTM} & 16.67 & 0.0591 & 59.82 & 68.00 & 16.60 & 0.0238 & 53.57 & 71.82 \\
        \textbf{+PT} & 17.53 & 0.0588 & 65.37 & 69.62 & 16.70 & 0.0235 & 54.15 & 72.41  \\
        \textbf{+FT} & 17.53 & 0.0596 & 66.44 & 69.80 & \textbf{14.10} & \textbf{0.0227} & \textbf{59.18} & \textbf{72.80} \\ \hline
        \textbf{HLW-LSTM} & 14.94 & 0.0586 & 64.73 & 69.74 & 16.00 & 0.0239 & 52.61 & 70.37 \\
        \textbf{+PT} & 15.27 & 0.0583 & 67.04 & 70.68 & 17.30 & 0.0239 & 53.69 & 71.21 \\
        \textbf{+FT} & \textbf{14.54} & \textbf{0.0537} & \textbf{69.41} & \textbf{72.18} & 17.00 & 0.0241 & 55.67 & 71.31 \\ \hline
    \end{tabular}

    \caption{Test results on RMSC-V2 and AAPD dataset. PT denotes the pre-training method. FT denotes the fine-tuned method on downstream task. OE and HL denote one-error and hamming loss respectively. ``(+)'' represents that higher scores are better and ``(-)'' represents that lower scores are better. ``-'' means results are not available.}
    \label{tab:results}
\end{table}

\subsection{Ablation Study}
In this section, some ablation studies to demonstrate the method proposed above are made on RMSC-V2 dataset. 

\paragraph{Encoder} To demonstrate that the label-wise encoder are better for multi-label classification, we introduce the traditional LSTM encoder and HAN encoder into pre-trainig mechanism. As shown in Table \ref{tab:lw-encoder}, LW-LSTM encoder has a huge improvemence than LSTM encoder for pre-training mechanism, and the same as HLW-LSTM. It indicates the label-wise approach is very effective and plays a vital role for MLTC task.
\begin{table}[htb]
    \centering
    \setlength{\abovecaptionskip}{0.3cm}
    \begin{tabular}{c|cccc}
        \hline
        \textbf{Model} & \textbf{OE(-)} & \textbf{HL(-)} & \textbf{Macro F1(+)} & \textbf{Micro F1(+)} \\ \hline
        LSTM+PT & 19.99 & 0.0623 & 48.36 & 63.71 \\
        \textbf{LW-LSTM+PT} & \textbf{17.53} & \textbf{0.0588} & \textbf{65.37} & \textbf{69.62} \\
        HAN+PT & 17.93 & 0.0623 & 50.02 & 64.94 \\
        \textbf{HLW-LSTM+PT} & \textbf{15.27} & \textbf{0.0583} & \textbf{67.04} & \textbf{70.68} \\
        \hline
    \end{tabular}
    \caption{Ablation tests on RMSC-V2 dataset with label-wise encoder.}
    \label{tab:lw-encoder}
\end{table}

\paragraph{Size of Candidates} To further explore the pre-training mechanism, we have tried several different size of candidates, such as 3, 4 and 5. As shown in Table \ref{tab:num}, three candidates in pre-training model with HLW-LSTM encoder and PT pre-training has better performance. Too many candidates may hurt the learning of documents. Too few candidates makes the task too simple to learn effectively.
\begin{table}
    \centering
    \setlength{\abovecaptionskip}{0.3cm}
    \begin{tabular}{c|cccc}
        \hline
        \textbf{n} & \textbf{OE(-)} & \textbf{HL(-)} & \textbf{Macro F1(+)} & \textbf{Micro F1(+)} \\ \hline
        \textbf{3} & \textbf{15.27} & \textbf{0.0583} & \textbf{67.04} & \textbf{70.68}\\
        4 & 15.80 & 0.0568 & 65.03 & 70.54 \\
        5 & 15.94 & 0.0577 & 65.23 & 70.46 \\
        \hline
    \end{tabular}
    \caption{Ablation tests on RMSC-V2 dataset with different size of candidates, denoted as n. HLW-LSTM+PT model is used.}
    \label{tab:num}
\end{table}

\subsection{Analysis}
To examine the ability of discovering the label correlations, we cite three correlated labels ``alternative'', ``metal'' and ``rock'' in music tagging task of RMSC-V2 dataset for example. In Table 5, for target song (i.e. document) ``Janes Addiction - Nothings Shocking'' possesses labels of ``alernative'' and ``rock''. We compute its top 50 similar songs via cosine similarity of document representation learnt respectively from three labels ``alternative'', ``metal'' and ``rock''. For each set of 50 similar songs, we calculate the label frequency (i.e., the proportion of songs that contain a specific label) and demonstrate the statistics in Table \ref{tab:label-prop}. The proportion smaller than 10\% are not presented.

From Table \ref{tab:label-prop} we can find that, for the target song, among its top 50 similar songs derived through ``alternative''-wise document representation, 76\% have label ``rock'', 42\% ``alternative'' and 26\% ``metal''. The similar observations can also be found for ``metal''-wise and ``rock''-wise document representation. This reveals that the proposed label-wise document representation is capable of capturing label correlations. Meanwhile, we can see all three columns has similar results, which means the label-wise representation on the original three labels are also similar. It is also interesting to notice that, even the target song doesn't hold the label ``metal'', it can also be accurately represented using the ``metal''-wise representation due to label ``metal'' is correlated with ``alternative'' and ``rock''. 

\begin{table}[htb]
    \centering
    \setlength{\abovecaptionskip}{0.3cm}
    \begin{tabular}{cc|cc|cc}
        \hline
         \multicolumn{2}{c|}{\textbf{Rep. on Alternative}} & \multicolumn{2}{c}{\textbf{Rep. on Rock}} & \multicolumn{2}{c|}{\textbf{Rep. on Metal}} \\ \hline
         \textbf{Label} & \textbf{P} & \textbf{Label} & \textbf{P} & \textbf{Label} & \textbf{P} \\ \hline
        rock & 76\% & rock & 86\% & rock & 82\% \\
        alternative & 42\% & metal & 38\% & metal & 50\% \\
        metal & 26\% & punk & 24\% & alternative & 24\% \\
        indie & 18\% & alternative & 22\% & punk & 12\% \\
        britpop & 18\% & indie & 18\% & indie & 10\% \\
        postpunk & 12\% & postpunk & 12\% & darkwave & 10\% \\
        punk & 12\% & ... & ... & ... & ... \\
        ... & ... & ... & ... & ... & ... \\ 
        \hline
    \end{tabular}
    \caption{Songs proportion over top 50 similar songs on ``Janes Addiction - Nothings Shocking''. P denotes the proportion of songs containing the list labels. Rep denotes the document representation on a specific label.}
    \label{tab:label-prop}
\end{table}

\begin{figure}[htb]
    \centering
    \subfigure[RMSC-V2]{
        \includegraphics[width=0.45\textwidth]{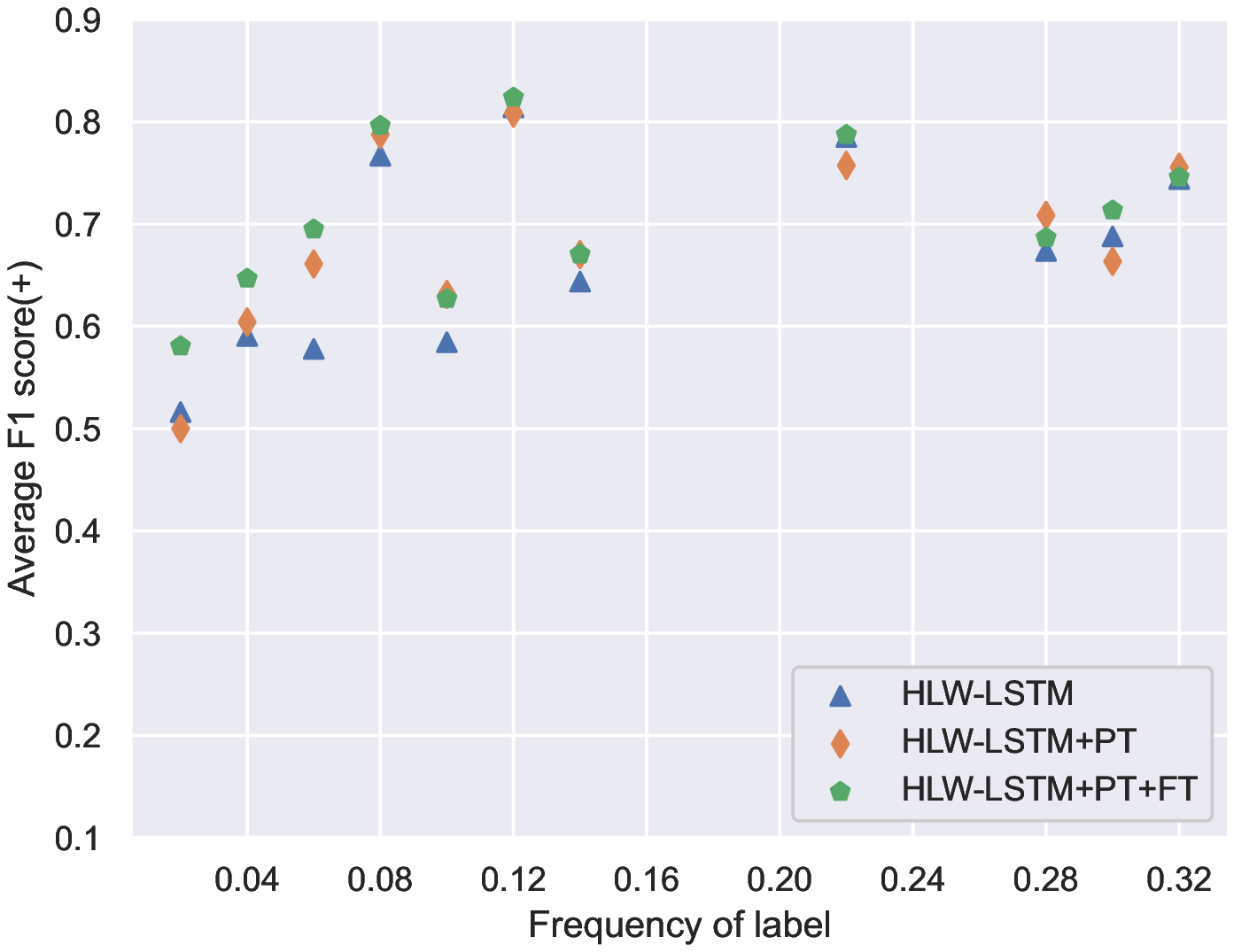}
    }
    \subfigure[AAPD]{
        \includegraphics[width=0.45\textwidth]{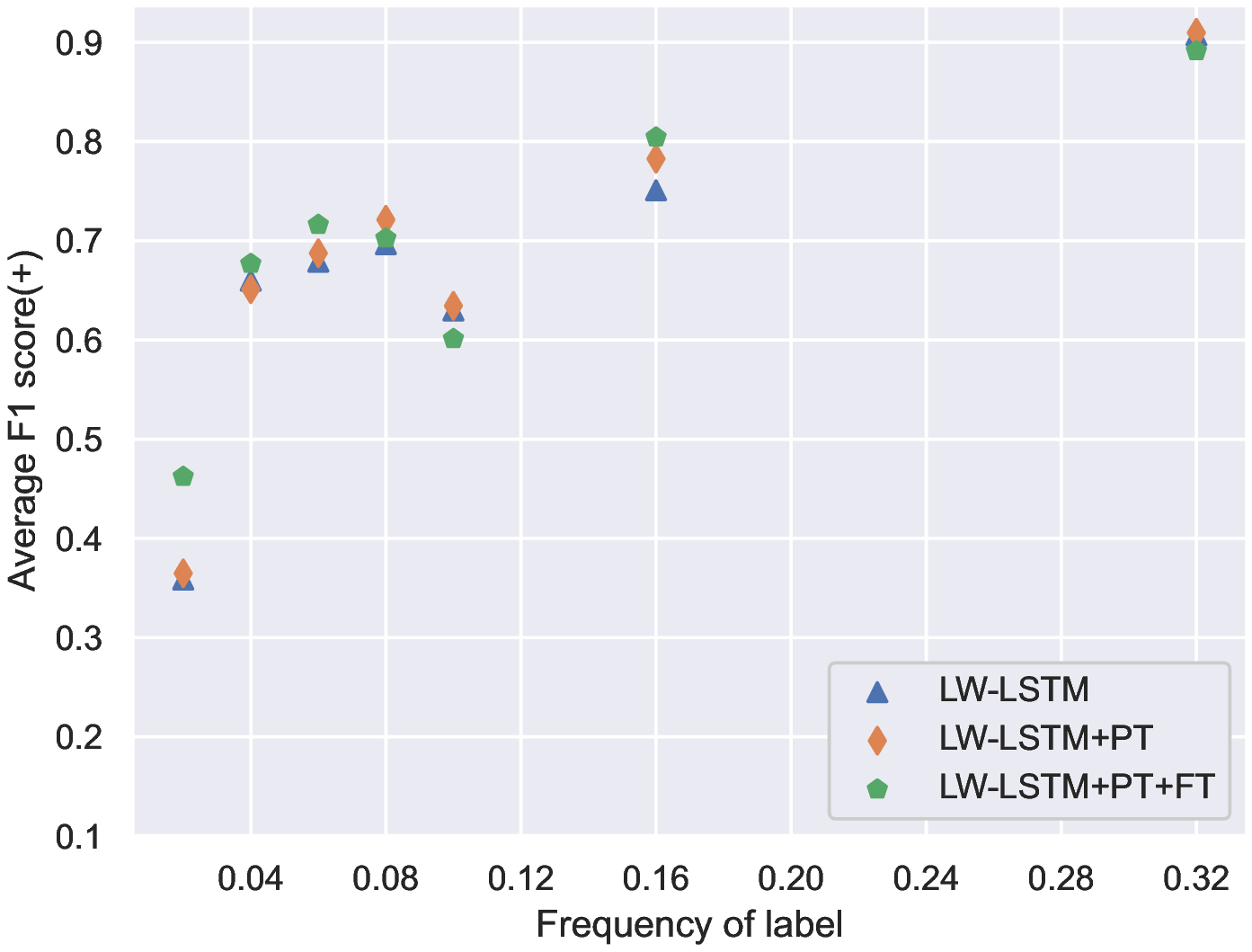}
    }
    \caption{Average F1 score with frequency of labels on RMSC-V2 and AAPD dataset.}
    \label{fig:rmsc-freq-f1}
\end{figure}

As mentioned above, our pre-training mechanism captures the label correlation. Thus, the labels have lower frequency is also predicted accurately. To demonstrate it, we count the frequency of each label and the corresponding F1 score. As shown in the Figure \ref{fig:rmsc-freq-f1}, frequency are discretized and average F1 score of labels on RMSC-V2 are calculated with HLW-LSTM encoder. And also LW-LSTM encoder for AAPD dataset. Obviously, the model have pre-training and fine-tuning has the best performance at most times and are better for lower frequency labels.

\subsection{Case Study}
We present some examples for HLW-LSTM models with or without pre-training and fine-tuning mechanism. As shown in Table \ref{tab:example}, five typical examples are selected from RMSC-V2 dataset in music tagging task. In the first three examples, models with pre-training accurately predict the whole label set. In the fourth, further using the fine-tuning mechanism obtains a accurate prediction. In the last, our pre-training and fine-tuning models both predict a extra label ``pop'' which is not appeared in the ground truth. However, we visit the website \footnote{\url{https://music.douban.com/subject/1774742/}} and find this song also have some labels similar to ``pop'' but not appear in the dataset. Note that the labels are made by human and maybe not accurate and complete. But for the label ``indie'' which is not predicted, this is where the model needs futher improvement.
\begin{table}[htb]
    \centering
    \setlength{\abovecaptionskip}{0.3cm}
    \begin{tabular}{l|lll}
        \hline
        \textbf{Ground Truth} & \textbf{HLW-LSTM} & \textbf{+PT} & \textbf{+PT+FT} \\ \hline
        jazz,soul & \textcolor{magenta}{jazz} & \textcolor{magenta}{jazz},soul & \textcolor{magenta}{jazz},soul \\
        classical,ost,piano & \textcolor{magenta}{ost,piano} & classical,\textcolor{magenta}{ost,piano} & classical,\textcolor{magenta}{ost,piano} \\
        folk,punk & \textcolor{magenta}{folk,punk},rock & \textcolor{magenta}{folk,punk} & \textcolor{magenta}{folk,punk} \\
        britpop,indie,rock & alternative,\textcolor{magenta}{britpop,indie,rock} & alternative,\textcolor{magenta}{britpop,indie,rock} & \textcolor{magenta}{britpop,indie,rock} \\
        electronic,indie & \textcolor{magenta}{electronic} & \textcolor{magenta}{electronic},pop & \textcolor{magenta}{electronic},pop \\
        \hline
    \end{tabular}
    \caption{Examples predicted by HLW-LSTM model with or without pre-training and fine-tuning mechanism on RMSC-V2 dataset. PT denotes the pre-training mechanism. FT denotes the fine-tuning procedure. Labels with red color means the intersection of three predicts.}
    \label{tab:example}
\end{table}

\section{Related Work}
There are many researches focus on Multi-Label Text Classification. In earlier years, traditional machine learning methods have been widely used in text classification, such as Naive Bayes, SVM, and so on. Therefore, researchers directly transform the multi-label task into several single-label tasks.

Binary relevance (BR) \cite{boutell2004learning} is the earliest method to learn several single-label classifier independently. Label Powerset (LP)\cite{tsoumakas2007multi} transforms it to a multi-class problem with one multi-class classifier trained on all unique label. Classifier chains (CC) \cite{read2011classifier} further convert it to a chain of single-label tasks. However, these methods cannot learn label-wise document representation and have very limited performance beacuse of insufficient learning.

In neural network models, sequence-to-sequence (Seq2Seq) framework can represent the labels correlation naturally. Therefore, CNN-RNN \cite{chen2017ensemble} and Sequence Generation Model (SGM)\cite{yang2018sgm} view the MLTC task as a sequence generation problem. Further, set-RNN\cite{qin2019adapting} presents an adaptation of RNN sequence models to the problem, where the target is a set of labels, not a sequence. However, these methods lack the dynamically modeling of label correlation through the fixed recurrent decoder. Significantly, the computation efficiency of Seq2Seq framework is a huge challenge for practical application.

Our label-wise document representation is very similar with word embedding. A word may have several meanings and a single static word embedding is difficult to completely represent the entire word. Thus, dynamic word embedding (i.e. pre-trained language models) is proposed to get a specific document representation with context, such as ELMo\cite{peters2018deep}, BERT\cite{devlin2018bert} and so on. But it's also static for different labels.

\section{Conclusions}
We propose a pre-training task and model LW-PT for multi-label text classification. Specifically, two label-wise encoders LW-LSTM and HLW-LSTM are introduced, which handle both short and long documents. In our method, label difference is modeled by the label-wise encoder. Label correlation is also captured in pre-training by the idea, that labels appeared together in a documents may have similar concepts. Experiments show that our method outperforms the previous approaches by a substantial margin. However, how to introduce unsupervised learning and transfer learning into label-wise pre-training mechanism is a further research in the future.

\section*{Acknowledgements}
The research is supported by the Fundamental Research Funds for the Central Universities.

\bibliographystyle{splncs04}
\bibliography{ref}

\end{document}